# MULTI-FACETED QUESTION COMPLEXITY ESTIMATION TARGETING TOPIC DOMAIN-SPECIFICITY

Sujay R[1], Suki Perumal[1], Yash Nagraj[1],
Anushka Ghei[2] and Srinivas K S[1]

[1]Department of CSE(AI & ML), PES University, Karnataka, India
[2]Department of CSE, PES University, Karnataka, India

## ABSTRACT

*Question difficulty estimation remains a multifaceted challenge in educational and assessment settings. Traditional approaches often focus on surface-level linguistic features or learner comprehension levels, neglecting the intricate interplay of factors contributing to question complexity. This paper presents a novel framework for domain-specific question difficulty estimation, leveraging a suite of NLP techniques and knowledge graph analysis. We introduce four key parameters: Topic Retrieval Cost, Topic Salience, Topic Coherence, and Topic Superficiality, each capturing a distinct facet of question complexity within a given subject domain. These parameters are operationalized through topic modelling, knowledge graph analysis, and information retrieval techniques. A model trained on these features demonstrates the efficacy of our approach in predicting question difficulty. By operationalizing these parameters, our framework offers a novel approach to question complexity estimation, paving the way for more effective question generation, assessment design, and adaptive learning systems across diverse academic disciplines.*

## 1. INTRODUCTION

The complexity of a Question is a multi-dimensional construct characterising the cognitive demand imposed by a given information retrieval request. This research paper explores this construct through a quantitative assessment of numerous factors that influence the efficiency with which the solver processes information, retrieves relevant knowledge, integrates disparate elements, and mobilises cognitive resources to derive an optimal solution. Traditionally, the determination of question difficulty has been approached as a pedagogical assessment, primarily focusing on a learner's surface-level comprehension required to answer the question, rather than considering the inherent complexity of the underlying knowledge or the cognitive processes required to generate abstract knowledge (*Lourdusamy et al., 2021*) [1].

However, this conventional methodology is limited as it does not dissect the multifaceted nature of the question's complexity. To overcome the limitations of traditional approaches to question calibration, we introduce a framework that explores a more comprehensive understanding of question difficulty. The parameters include Topic Salience, Topic Retrieval Index, Topic Coherence Score, and Topic Superficiality. Utilising these parameters, we can estimate the complexity of a question. The major performance improvement of our proposed model is predominantly observed in the context of complex questions. This is primarily due to the model's robustness in managing the cognitive demand imposed by a given information retrieval request, even as the number of information components increases.

This attribute is particularly beneficial in university-level subject domains including but not limited to information retrieval and knowledge management, where the complexity of queries can vary significantly. Thus, our model's ability to maintain high-performance levels, especially with complex questions, underscores its potential to enhance the effectiveness of systems in these domains. Moreover, our parameters are specifically designed to account for the textbook-specific difficulty of questions. This ensures that our model is finely tuned to the academic context, further enhancing its applicability and effectiveness.

## 2. RELATED WORK

Various approaches to classifying responses in question-answering (QA) systems often involve constructing taxonomies for question categorization, employing either flat (*David et al., 2000*)[2]. or hierarchical (*Takaki et al., 2000*)[3]. grouping schemas. Additionally, domain-specific taxonomies have been developed to handle particular question varieties, such as those related to opinions (*Yang-woo Kim et al., 2014*) [4], data sources, analysis types, and response formats (*Bayoudhi et al., 2013*) [5]. Similarly, the utilisation of Bloom's Taxonomy primarily captures the basic learner's comprehension level, rather than the need for deeply processed knowledge that can be abstracted and transferred across contexts (*Sabine Ullrich et al., 2021*)[6]. Our proposed method surpasses existing approaches by moving beyond the semantics of the question and learner comprehension levels and instead delves into the inherent difficulty of the subject matter and the mental effort needed to find the answer, leading to a more nuanced understanding of question difficulty.

The study by *Hoerl et al., (2021)*[7] examines how student characteristics and cognitive skills, differing levels of text complexity, and reading comprehension question types affect different types of reading outcomes. However, this study does not explore how different aspects of text complexity (such as cohesion, and decoding) individually contribute to these outcomes. The study also identifies critical and inferential questions as the most difficult types of comprehension questions without taking into consideration why these types are more challenging. Several papers use multiple-choice reading comprehension questions (*Zhang et al., 2023*)[8], free recall (*Taconnat et al.,2023*) [9] and oral reading fluency (*Jungjohann et al., 2023*)[10] as measures of reading outcomes. However, these measures may not fully capture the complexity of reading comprehension.

In the paper (*Y. Zhang et al., 2018*)[11], the evaluation of the model prioritises generalisation metrics over domain-specific context, often neglecting the depth of knowledge extraction within specific subject areas in favour of broader applicability. Our model's efficacy lies in discovering highly specific latent topics within a subject domain. *Seyler et al., (2016)*[12] have researched the methods of generating knowledge questions from knowledge graphs, but do not take into consideration the importance of the topic relative to the subject-domain when evaluating question difficulty. This limits the applicability of their findings in an educational context, where the relevance of a topic to the syllabus is a crucial factor in determining its importance. In our approach, we incorporate a weighted topic importance mechanism, which aligns closely with the syllabus of the subject domain, ensuring that the extracted features directly address the learning outcomes outlined in the syllabus.

## 3. METHODOLOGY

Our architecture conducts a multi-facet assessment of questions to assess their complexity, with a focus on incorporating knowledge domain-specific features. The difficulty index of a given question is estimated with the set of computed parameters: Topic Retrieval Cost, Topic Salience, Topic Coherence Score, and Topic Superficiality. Each parameter contributes a distinct and informative dimension or facet, thereby mitigating potential biases or constraints inherent in other parameters. As a result, this provides an exhaustive portrayal of the attributes outlining question complexity.

### 3.1. Topic Retrieval Cost

This metric, denoted as "ρ", quantifies the thematic scarcity of a specific topic(T) within a document corpus associated with a particular subject domain. It is a definitive measure of a topic's relative prominence and coverage within the subject matter capturing the information density associated with T relative to the overall thematic landscape of C. We examine $\rho(T, C)$, proposing that the prevalence of a topic(T) within the corpus(C) directly impacts the difficulty in recalling or retrieving information about that topic as demonstrated by (*Kubik et al., 2021*) [13]. Owing to the sparse topical representation within the subject materials, queries concerning topics with limited exposure lead to increased difficulty in retrieving relevant knowledge. We further integrate results of *Lei et al., (2023)* [14] regarding information retrieval research. The computation of this metric leverages a topic model trained on the textbook corpus. To evaluate retrieval efficacy, we calculate the percentile rank of the target topic against a distribution table from our pre-trained topic model (refer to 4.1.2).

$$\rho(T, C) = 1 - \frac{r(T, C)}{|C|}$$

where,

- r(T, C) is the rank or ordinality of topic T within the set of topics in corpus C,
- |C| is the cardinality (size) of the set of topics in corpus C

This formulation represents the Topic Retrieval Index as a normalised value between 0 and 1, where:

- $\rho(T, C) = 1$ indicates that topic T is a highly prevalent topic in corpus C (r(T, C) = 1)
- $\rho(T, C) = 0$ indicates that topic T is a scarcely prevalent topic in corpus C (r(T, C) = |C|)

### 3.2. Topic Salience Score

The Topic Salience Score (η) estimates question difficulty based on the concept of entity centrality where the entity (e) corresponds to the topic (T) of the question; as explored by (*Yuling et al., 2020*) [15] in their topic popularity definition. We use Knowledge graphs (KG) that consist of ontologies for each topic (*Kejriwal et al., 2022*) [16] The presence of a link between two topics in the Knowledge Graph indicates that the linked article contains foundational knowledge, or related concepts that are relevant to comprehending the subject matter of the current topic. To compute the Topic Salience Score, the article(A) corresponding to the topic T in the KG is obtained. We analyse the inbound links associated with the KG corresponding to the article A. These links serve as indicators of how extensively the topic 'T' is referenced within the KG. We retrieve a set, denoted as **Γ**, consisting of the incoming links for article A. We also obtain the number of incoming links for each article within **Γ**. The Salience(η) is then computed by dividing the number of the incoming links for A by the average number of incoming links for each article within the **Γ** set.

$$\eta = f(C, C_\Gamma) = \frac{C}{(\Sigma C_\Gamma / n)}$$

where,

- C is the number of incoming links to the Wikipedia article corresponding to the topic.
- $C_\Gamma$ is the list of counts of incoming links for each article within the $\Gamma$ set.
- $\Sigma C_\Gamma$ is the sum of all counts in the list $C_\Gamma$.
- n is the number of elements in the list $C_\Gamma$.

A higher salience score suggests the topic is a central concept with numerous connections within the subject domain, potentially indicating a more fundamental topic. Conversely, a lower salience score implies a less frequently referenced topic, potentially signifying a more specialised or intricate concept. However, the friendship paradox; (*Sidorov et al., 2021*) [17], can affect the perceived difficulty of questions based on the Topic Salience Score by inflating ($\Sigma C_\Gamma / n$).

Since, topics linked to central topics will tend to have much more links than the current topic, altering the score by overestimating the relative centrality of the target topic compared to its linked articles. To mitigate this phenomenon, our approach incorporates alternate parameters (mentioned in subsequent sections) designed to address its effects.

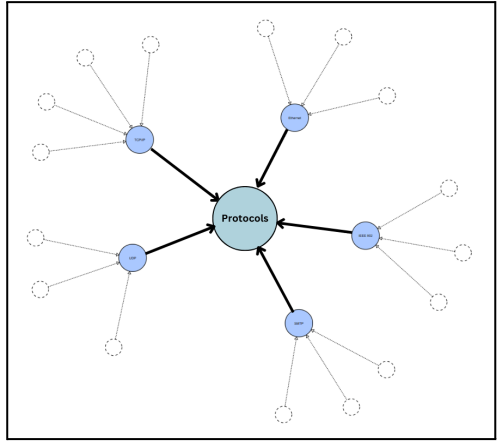

Figure 1: Graphical Representation of Topic Salience

### 3.3. **Topic Coherence Score**

This metric, denoted by λ, quantifies how closely related two topics are by capturing their tendency to appear together in the KG. The underlying assumption is that questions that require reasoning about closely related topics are generally easier to answer than those that require reasoning about distant topics. Building on recent advancements in coherence measures for evaluating relationships between complex word subsets (*Rahimi et al., 2023*) [18], we have refined our score to better capture the depth of topic relationships. Formally, λ(t1, t2) measures the conditional probability of observing topic t2 given the presence of topic t1. We capture coherence by leveraging the link structure of the KG, which mirrors the relationships and semantic associations among topics. Specifically, the coherence between two topics t1 and t2, is defined as the Jaccard coefficient, measuring the overlap between these two topics.

$$\lambda(t1, t2) = \frac{|L(t1) \cap L(t2)|}{|L(t1) \cup L(t2)|}$$

where,

- t1 and t2 represent the two topics
- L(t) represents the set of incoming links for a topic (t)

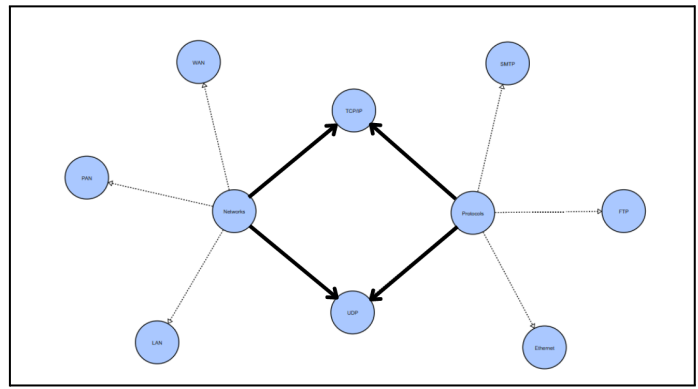

Figure 2: Graphical Representation of Topic Coherence

### 3.4. **Topic Superficiality Index**

The Topic Superficiality Index (**Ω**) serves as a quantitative measure assessing the vertical depth within a hierarchical knowledge structure (*Peng et al., 2023*) [19]. It is derived by calculating the topological distance between a given topic node and the primary subject node, which acts as the origin of the sub-knowledge graph. The greater depth of a topic node in the hierarchy corresponds to the higher superficiality of the given topic. Mathematically, the Topic Superficiality Index (**Ω**) can be represented as the number of edges required to navigate from the root node to the topic node in the question. Deeper topics in a Knowledge graph require traversing through more specific and less general concepts. A question that requires understanding these deeper concepts likely requires a stronger foundation in the subject matter. The Topic Superficiality Index (**Ω**) is particularly useful for assessing the difficulty of questions that require a deep understanding of a specific sub-topic or niche area.

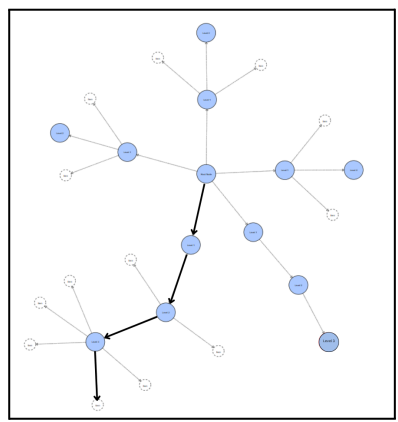

Figure 3: Graphical Representation of Topic Superficiality Index

## 4. **EXPERIMENTAL SETUP**

The experimental setup consists of four main phases: question generation, preprocessing, knowledge graph nomination, and experimental evaluation.

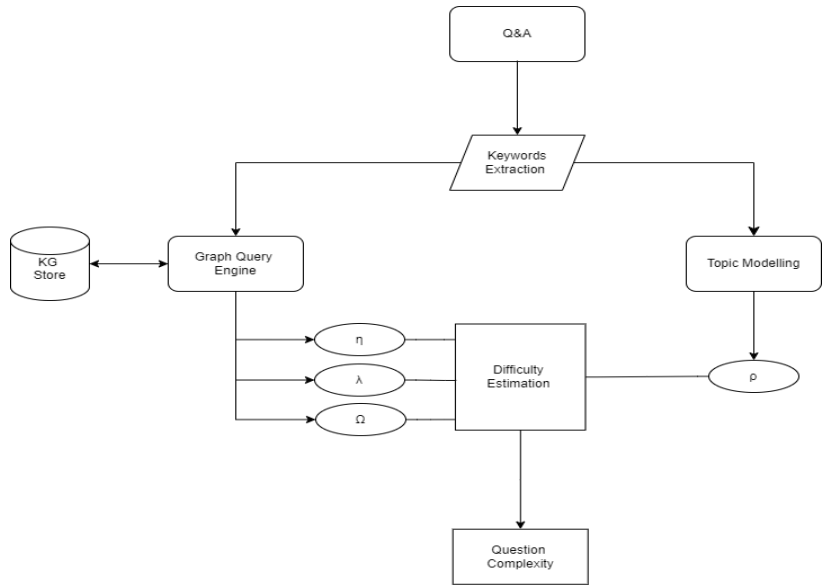

Figure 4: Experimental Framework

## 4.1. Question Generation

This phase involves the Generation of Questions from a predetermined set of domain-specific corpora. We are focusing on the subject of Computer Networks, utilising our university's prescribed: *"Computer Networking: A Top- Down Approach", James F. Kurose, Keith W(2017)* as the contextual source for the subject matter. Chapters and sections are extracted using the pdfminer library, facilitating the segmentation of the text into semantically coherent units.

For each demarcated section, GPT-4 is leveraged to generate a set of 100 questions, contingent upon the corresponding textual sections (*Chevalier et al., 2024*) [20]. To ensure heterogeneity within the question set, multiple prompts with varying topic requirements are employed during the question generation process: single-topic(closed-ended), multi-topic(open-ended), memory-based and analytical (*Olney et al., 2023*)[21], (*Amyeen et al., 2023*) [22]. This process facilitates the formulation of a diverse array of question templates, resulting in questions that are knowledge-intensive. Followed by an item de-duplication procedure wherein redundant questions are identified and subsequently eliminated. This is achieved by obtaining the embeddings of all questions using `text-embedding-3-large` from OpenAI, if the cosine similarity of any two questions exceeds `0.97`, the duplicate is removed, ensuring the uniqueness of the generations (*Murugan et al., 2022*) [23].

### 4.1.1.  Preprocessing

We characterise the multifaceted nature of question difficulty by extracting pertinent features from the questions. KeyBERT is employed to extract keywords from the question, and its corresponding answer (*Boshko et al., 2022*)[24]. These keywords serve as the foundation for computing four intricate difficulty metrics: Topic Salience Score,  Topic Retrieval Cost, Topic coherence, and Topic Superficiality.

Table 1: Sample Question Analysis and Keyword Extraction

| Question Template | Example Question | Extracted Keywords |
|---|---|---|
| Single-Topic | Explain the significance of Address Resolution Protocol (ARP) in local area networks and discuss a scenario where understanding ARP is crucial for successful communication between hosts on different subnets. | [ 'arp', 'subnets', 'communication' ] |
| Multi-Topic | How can the cryptographic principles of confidentiality, message integrity, and end-point authentication be applied to enhance the security of financial transactions in online banking systems? Discuss the potential vulnerabilities that attackers might exploit in such systems and propose specific cryptographic techniques or protocols that could mitigate these risks effectively. Provide a detailed explanation supported by relevant examples to demonstrate the real-world implications. | [ 'banking', 'authentication','cryptographic', 'security', ] |
| Memory Based | Explain how the transition from traditional hierarchical data centre network designs to highly interconnected topologies, such as fully connected architectures, can address the limitations of host-to-host capacity in large-scale data centres. Analyse the implications of this shift on network performance, scalability, and flexibility within the context of modern cloud applications. | [ 'cloud', 'scalability', 'topologies', 'data centre' ] |
| Analytical | Assuming you are tasked with designing a network application that requires real-time data transmission with low latency, high reliability, and minimal overhead. Discuss and justify whether you would choose to implement this application using TCP or UDP as the underlying transport protocol. | [ 'udp', 'tcp', 'latency', 'overhead', 'congestion' ] |

### 4.1.2. Topic Retrieval Cost

The Topic Retrieval Cost, a domain-specific metric, quantifies the scarcity or prevalence of a question's topic within the subject corpus. It captures the information density associated with the topic relative to the overall domain, with the premise that topics receiving limited topical coverage impose higher difficulty in retrieving relevant knowledge.

**4.1.2.1. Pre-Training**

To calculate Retrieval Cost, we are training BERTopic on multiple textbooks for the subject. BERTopic's robust architecture (Grootendorst et al., 2022)[25] and adaptability make it well-suited for extracting thematic structures from our educational text corpus. We leverage BERT for document embedding to capture the semantic meaning and contextual relationships crucial for understanding conceptual connections within educational materials. UMAP is then employed for dimensionality reduction, preserving inter-document relationships for accurate clustering. The reduced embeddings are clustered using HDBSCAN which handles noise and clusters of varying densities, ensuring cohesive and representative topics. After BERTopic is trained on the corpus, a topic distribution table is obtained containing each topic and its document frequency in the training text.

**4.1.2.2. Topic Model Inference**

For a given set of keywords extracted from a question. Each keyword undergoes processing through BERTopic. Due to BERTopic's limitations in handling brief keywords, we transform each keyword into a concise description before inputting it to BERTopic. The resulting topic model assigns a topic cluster to the transformed keyword. We then analyse the placement of this topic within the topic distribution table obtained from pre-training and compute the Retrieval Cost using the methodology outlined in 3.1. Given the presence of multiple keywords per question, we aggregate metrics such as the minimum, maximum, mean, and sum across all keywords within a question.

Example:
**Question 1: $\rho_\mu = 0.72$**
*Explain the self-scalability of a Peer-to-Peer (P2P) architecture in file distribution compared to a client-server architecture. Discuss the factors influencing distribution time, the minimum distribution time formulas for both architectures and how the P2P architecture's design allows for efficient file sharing.*

**Question 2: $\rho_\mu = 0.88$**
*Explain the vulnerabilities associated with the Wired Equivalent Privacy (WEP) security protocol in IEEE 802.11 wireless networks, and discuss how the adoption of the 802.11i standard in 2004 addressed these weaknesses.*

In this example, Question 1 explores the niche application of P2P architecture. The lower $\rho_\mu$ signifies the topic's limited coverage within the textbook. Conversely, Question 2, addresses the fundamental concept of IEEE 802.11 wireless protocol. The higher $\rho_\mu$ reflects the topic's greater presence within the textbook materials.

**4.1.3. Knowledge Graph Nomination**

To calculate Topic Salience, Topic coherence and Topic Superficiality we choose DBpedia, a structured interface to Wikipedia, We use the hyperlink structure present in Wikipedia articles, making it suitable for analysing inbound and outbound links between articles.

#### 4.1.4.  Topic Salience

When provided with a keyword, we query the DBpedia database. The query retrieves a set of Wikipedia articles, denoted as **Γ**, that contain incoming links pointing to the specific Wikipedia article associated with the given keyword. In addition to obtaining this set of articles (**Γ**), the query also fetches the count of incoming links for each article within the **Γ** set. Then topic salience is calculated by the formula given in the methodology section 3.2. Similar to the topic retrieval cost calculation, we take aggregate metrics including minimum, maximum, mean, and sum of the topic salience scores across all keywords in a given question.

Example:
**Question 1: $\eta_\mu$ = 2.63**
*Explain the significance of BGP in the context of Internet routing, detailing its role in inter-Autonomous-System communication. Analyse how BGP's route-selection algorithm navigates the complex network landscape.*

**Question 2: $\eta_\mu$ = 6.54**
*In the context of TCP connections, explain how sequence numbers and acknowledgement numbers work together to ensure reliable data transfer. Describe the potential challenges that can arise if sequence and acknowledgement numbers are not properly utilised, and propose strategies to mitigate these challenges effectively.*

The examples showcase the relationship between Topic Salience($\eta$) and perceived question difficulty. Question 1, exploring the specialised BGP routing, exhibits a lower $\eta_\mu$, suggesting a niche topic within networking. Conversely, Question 2 on fundamental TCP mechanisms exhibits a higher $\eta_\mu$, reflecting its centrality and interconnectedness within computer networking, potentially indicating a less challenging question due to broader coverage and established connections.

#### 4.1.5.  Topic Coherence

To compute the Topic Coherence for each keyword extracted from a question, we query the DBpedia database to retrieve the set of incoming links (Γ) that point to the corresponding Wikipedia articles. For every pair of keywords k1 and k2 within the question we then calculate the Jaccard similarity coefficient between the sets of incoming links Γ(k1) and Γ(k2). To obtain the final Topic Coherence feature for a given question, we aggregate the pairwise coherence scores by computing the minimum, maximum, mean, and sum across all keyword pairs.

Example:
**Question 1: $\lambda\mu$= 0.317**
Explain the significance of choosing between TCP and UDP when developing a client-server application. Consider the implications of selecting each protocol on the reliability, delivery guarantees, and performance of the application.

**Question 2: $\lambda\mu$ = 0.00413**
Analyse the technical shortcomings of ARP that prevent it from being suitable for WAN routing. Explore how WANs achieve communication without direct ARP reliance.

Question 1, refers to TCP and UDP for client-server applications, exhibiting higher coherence. This is likely due to the frequent co-occurrence of TCP and UDP in discussions related to network protocols. Question 2 is contrasting ARP with WAN routing, and presents a lower mean coherence. This suggests a weaker association, as ARP is typically discussed in the context of LANs rather than WANs, potentially leading to a more challenging cognitive process for integrating these concepts.

### 4.1.6. Topic Superficiality

To assess the thematic relevance of each keyword, we compute a "superficiality score" based on its proximity within the Wikipedia network to the central topic of "Computer Networks." For each keyword, we identify the corresponding Wikipedia article (A1) and measure its graph distance to the "Computer_network" article (A'). This distance, representing the number of intermediary nodes in the shortest path between A' and A1, serves as the superficiality score. By analysing the distribution of these scores across all keywords within a question, we derive aggregate metrics including the minimum, maximum, mean, standard deviation, and sum.

### Question 1: $\Omega_\mu = 2$
*Explain the significance of the Simple Mail Transfer Protocol (SMTP) in the context of Internet mail communication. How does SMTP facilitate the reliable transfer of email messages from a sender to a recipient?*

### Question 2: $\Omega_\mu = 1.2$
*Explain the vulnerabilities associated with the Wired Equivalent Privacy (WEP) security protocol in IEEE 802.11 wireless networks, and discuss how the adoption of the 802.11i standard in 2004 addressed these weaknesses.*

In this example, Question 1, focusing on the Simple Mail Transfer Protocol (SMTP), exhibits a higher mean superficiality score, suggesting that SMTP represents a more specialised concept within the broader domain of computer networks. This implies that understanding SMTP might require navigating through more fundamental concepts before reaching this specific protocol. Question 2, addressing the vulnerabilities of the WEP security protocol in IEEE 802.11 wireless networks, presents a lower mean superficiality score. This indicates that WEP is a more foundational concept within the domain, potentially requiring less prerequisite knowledge and thus posing a lesser challenge in terms of topic depth. Finally, for each question, a comprehensive statistical summary is derived, encompassing various aggregates of these features, thereby encapsulating these features into one vector for evaluation.

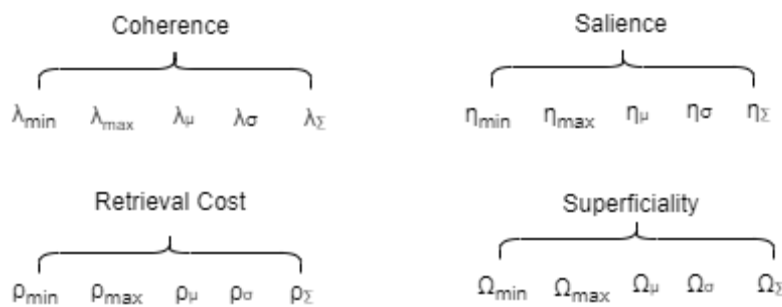

Figure 5:  Statistical Summary of Evaluation Metrics

## 4.2. Experimental Evaluation

To ensure the acquisition of accurate and detailed difficulty labels, the annotation process utilises the ELO ranking system (*Cheng et al., 2024)[26]*. This system facilitates pairwise comparisons, wherein the evaluators compare pairs of questions and select the one they perceive as exhibiting a higher degree of difficulty. The iterative nature of this process results in a ranked leaderboard of question complexity, reflecting the collective judgement of the evaluators. Each question is subsequently assigned a difficulty label in accordance with its percentile ranking within this leaderboard, providing a continuous measure of difficulty that captures the relative complexity of each question within the dataset.

To assess the difficulty of our set of 100 questions, we employ human evaluators to evaluate a subset of the dataset and provide their reasoning. This evaluated subset is used in a few-shot prompting an LLM agent powered by GPT-4. The agent proceeds to evaluate the remaining set. Leveraging the ELO ranking system's efficiency, pairwise comparisons only require a limited subset of questions, rather than an exhaustive evaluation of every question against all others.

Given the exploratory nature of this study and the primary focus on establishing the efficacy of the proposed feature set, we opt for a simple yet robust regression model – Linear Regression. This choice is motivated by its interpretability, allowing for a clear understanding of the relationship between the extracted features and predicted question difficulty. The model is trained using the curated dataset, and its performance is evaluated through 5-fold cross validation. Mean squared error (MSE) and R-squared serve as the primary evaluation metrics to assess the model's predictive accuracy and generalizability.

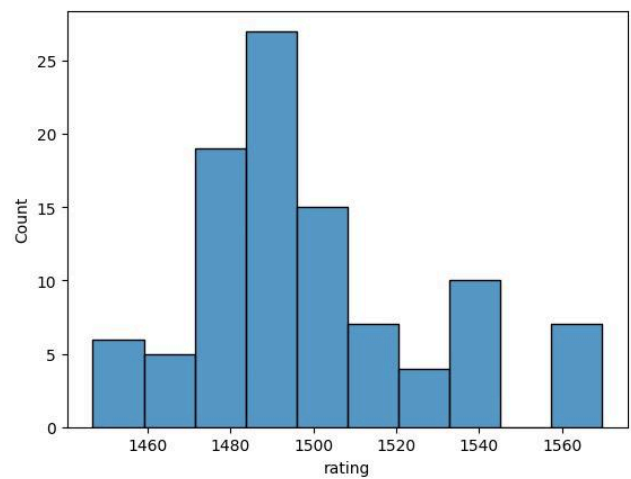

Figure.6: Distribution of Elo rankings in the curated dataset (k=20)

## 5. RESULTS

The Linear Regression model trained on the extracted features achieved a mean squared error (MSE) of 3.06 and an R-squared value of 0.42. While the model demonstrates the ability to capture a significant portion of the variance in question difficulty, there remains room for improvement in predictive accuracy. The moderate R-squared value suggests that more sophisticated deep learning techniques could potentially enhance the model's performance. Further analysis of the model's errors and the relative importance of different features will provide valuable insights for future refinements.

## 6. DISCUSSION

The proposed approach contributes to the enhancement of question generation and assessment in various educational and evaluative contexts, enabling more accurate estimation of question difficulty and potentially facilitating the development of adaptive learning systems, personalised assessments, and improved curriculum design. However, Human annotation is susceptible to errors arising from several factors such as subjectivity, bias, ambiguous annotation guidelines, limited domain expertise, data intricacies, and class imbalance (*Kurdi et al., 2020) [27]*. These factors can result in inconsistencies and biassed annotations, which in turn, can adversely affect model estimation and performance. To address these challenges, implementing quality assurance procedures, employing multiple annotations with adjudication, utilising active learning techniques, iteratively refining the process, and leveraging automated annotation assistance can enhance annotation quality and mitigate the potential impact of errors in model estimation.

When utilising DBpedia**,** a knowledge base derived from Wikipedia articles, there is a potential issue in accurately determining the true depth of topic coverage due to the vast number of hyperlinks present within the articles. While Wikipedia articles often contain numerous hyperlinks to other articles, leading to a hypergraph with a high index of links between articles, this proliferation of hyperlinks can induce a perception of depth and interconnectivity, irrespective of the articles' potentially surface-level treatment of the corresponding topics. To better assess true topic depth, it may be more beneficial to utilise a knowledge graph specifically designed to represent subject relations with greater clarity and depth.

Assessing the abstraction level and the regurgitative nature of questions could provide valuable insights into question complexity *(Das et al., 2020)* [28]. Questions involving highly abstract concepts and requiring synthesis or application of knowledge, tend to be cognitively demanding. Incorporating these parameters into the existing framework could offer a more nuanced understanding of question complexity, capturing both the conceptual abstraction and the depth of cognitive processing required.

## 7. CONCLUSION

The proposed multi-faceted question complexity estimation framework represents a notable advancement in the field of Computational Linguistics in Education, offering a comprehensive and domain-aware approach to accurately estimate the inherent difficulty of questions. The key innovation lies in the formulation of four distinct parameters - **Topic Retrieval Cost, Topic Salience, Topic Coherence Score, and Topic Superficiality Index.** Collectively, these metrics capture the inherent challenges posed by a question, ranging from the scarcity of topic coverage in the subject materials to the cognitive demands of integrating disparate concepts. By accounting for the nuanced interplay of factors, this work paves the way for more tailored and adaptive question generation and evaluation processes, ultimately improving the quality of instruction and evaluation across various academic disciplines. In conclusion, we have successfully translated our conceptual definition of question difficulty, as outlined in the Introduction, into tangible and quantifiable metrics, thereby developing a novel construct of question complexity through a systematic methodology.

# ACKNOWLEDGEMENTS

The authors would like to express their acknowledgement for the financial support provided by PES University.